\documentclass{sn-jnl}

\PassOptionsToPackage{round}{natbib}

\usepackage{hyperref}
\usepackage[dvipsnames]{xcolor}
\usepackage[utf8]{inputenc} 
\usepackage[T1]{fontenc}    
\usepackage{url}            
\usepackage{booktabs}       
\usepackage{amsmath}
\usepackage{amsfonts}       
\usepackage{nicefrac}       
\usepackage{microtype}      
\usepackage{multirow}
\usepackage{graphicx}
\usepackage{booktabs}
\usepackage{geometry}
\usepackage{subcaption}
\usepackage{caption}
\usepackage{natbib}
\usepackage{amssymb} 
\usepackage{pifont}  
\usepackage{array}    
\usepackage{makecell} 
\usepackage{adjustbox}
\usepackage{cleveref}
\usepackage{tablefootnote}

\hypersetup{
   breaklinks=true,   
   colorlinks=true,  
   linkcolor=Bittersweet,
   citecolor=Periwinkle,
   urlcolor=Bittersweet
}
\definecolor{red}{HTML}{ca0020}
\definecolor{lightred}{HTML}{f4a582}
\definecolor{lightblue}{HTML}{92c5de}
\definecolor{green}{HTML}{008837}
\definecolor{blue}{HTML}{2c7bb6}

\title{AlphaNet: Scaling Up Local-frame-based Neural Network Interatomic Potentials}

\author[1]{\fnm{Bangchen} \sur{Yin}}\email{yinbc24@mails.tsinghua.edu.cn}
\equalcont{These authors contributed equally to this work.}

\author*[2]{\fnm{Jiaao} \sur{Wang}}\email{wangjiaao0720@utexas.edu}
\equalcont{These authors contributed equally to this work.}

\author[5]{\fnm{Weitao} \sur{Du}}\email{duweitao.dwt@alibaba-inc.com}

\author[6]{\fnm{Pengbo} \sur{Wang}}\email{Wang.Pb@outlook.com}

\author[7]{\fnm{Penghua} \sur{Ying}}\email{hityingph@tauex.tau.ac.il}

\author[4]{\fnm{Haojun} \sur{Jia}}\email{haojunjia@deepprinciple.com}

\author[8, 9]{\fnm{Zisheng} \sur{Zhang}}\email{zishengz@stanford.edu}

\author*[3]{\fnm{Yuanqi} \sur{Du}}\email{yd392@cornell.edu}

\author[3]{\fnm{Carla} \sur{Gomes}}\email{gomes@cs.cornell.edu}

\author*[4]{\fnm{Chenru} \sur{Duan}}\email{duanchenru@deepprinciple.com}

\author*[2]{\fnm{Graeme} \sur{Henkelman}}\email{henkelman@utexas.edu}

\author*[1]{\fnm{Hai} \sur{Xiao}}\email{haixiao@tsinghua.edu.cn}

\affil[1]{\orgdiv{Department of Chemistry}, 
          \orgname{Tsinghua University}, 
          \orgaddress{\city{Beijing}, \postcode{100084}, \country{China}}}

\affil[2]{\orgdiv{Department of Chemistry and the Oden Institute for Computational Engineering and Sciences}, 
         \orgname{The University of Texas at Austin}, 
         \orgaddress{\city{Austin}, \state{TX}, \postcode{78712-0165}, \country{USA}}}

\affil[3]{\orgdiv{Department of Computer Science}, 
          \orgname{Cornell University}, 
          \orgaddress{\city{Ithaca}, \state{NY}, \postcode{14850}, \country{USA}}}

\affil[4]{\orgdiv{Deep Principle, Inc.}, 
          \orgaddress{\city{Cambridge}, \state{MA}, \postcode{02139}, \country{USA}}}
          
\affil[5]{ 
         \orgname{DAMO Academy}, 
         \orgaddress{\city{Beijing}, \postcode{100020}, \country{China}}}

\affil[6]{\orgdiv{Department of Chemical and Biological Engineering}, 
         \orgname{The Hong Kong University of Science and Technology}, 
         \orgaddress{\city{Hong Kong}, \postcode{999077}, \country{China}}}

\affil[7]{\orgdiv{Department of Physical Chemistry}, 
         \orgname{Tel Aviv University}, 
         \orgaddress{\city{Tel Aviv}, \postcode{6997801}, \country{Israel}}}

\affil[8]{\orgdiv{Department of Chemical Engineering}, 
         \orgname{Stanford University}, 
         \orgaddress{\city{Stanford}, \state{CA}, \postcode{94305}, \country{USA}}}

\affil[9]{\orgdiv{SLAC National Accelerator Laboratory}, 
         \orgaddress{\city{Menlo Park}, \state{CA}, \postcode{94025}, \country{USA}}}

\begin{document}

\maketitle

\begin{abstract}

Molecular dynamics simulations demand an unprecedented combination of accuracy and scalability to tackle grand challenges in catalysis and materials design. To bridge this gap, we present AlphaNet, a local-frame-based equivariant model that simultaneously improves computational efficiency and predictive precision for interatomic interactions. By constructing equivariant local frames with learnable geometric transitions, AlphaNet encodes atomic environments with enhanced representational capacity, achieving state-of-the-art accuracy in energy and force predictions. Extensive benchmarks on large-scale datasets spanning molecular reactions, crystal stability, and surface catalysis  (Matbench Discovery and OC2M) demonstrate its superior performance over existing neural network interatomic potentials while ensuring scalability across diverse system sizes with varying types of interatomic interactions. The synergy of accuracy, efficiency, and transferability positions AlphaNet as a transformative tool for modeling multiscale phenomena, decoding dynamics in catalysis and functional interfaces, with direct implications for accelerating the discovery of complex molecular systems and functional materials. Our code and data are available at \url{https://github.com/zmyybc/AlphaNet}.

\end{abstract}

\section{Introduction}

Molecular dynamics (MD) simulations have become essential for investigating and elucidating complex phenomena across diverse fields, including biology, catalysis, and energy engineering~\citep{Fermi, Fermi1, alder,alder1, mdreview,md1,md2,md3}. While atomic forces necessary for MD simulations can be derived from quantum mechanical approaches such as density functional theory (DFT)~\citep{aimd}, the computational cost associated with such first-principles methods severely limits their applications to small-sized systems ($\sim{}10^3$ atoms) and short timescales ($\sim{}10^1$ ps). Consequently, a wide spectrum of phenomena occurring over much larger scales remain inaccessible, even with the most powerful supercomputers.

Classical force fields, which rely on predefined mathematical forms, offer a computationally efficient alternative, allowing simulations of larger systems over extended trajectories~\citep{lj,amber,charmm, gromos, reaxff, Karplus2002}. However, the simplicity of these models often compromises their accuracy, necessitating a delicate trade-off between computational efficiency and the fidelity of the simulated dynamics, especially for reactive chemistry~\citep{reaxff, tersoff, iftimie2005ab,zhao2025harnessing}.

The advent of machine learning techniques has introduced a promising solution to this dilemma. By training models on data derived from first-principles calculations, neural network interatomic potentials (NNIPs) can potentially achieve the accuracy of first-principles methods while maintaining the computational efficiency of classical force fields~\citep{mace,nequip,dpa2,review1,review2,review3,review4}. 
Unlike traditional models that rely on explicit functional forms to describe the bonded and non-bonded interactions, NNIPs are more flexible, learning to predict interactions based on the positions of atoms and their chemical identities~\citep{bp}.

As interest in applying NNIPs to large-scale atomistic systems continues to grow, the field has seen a surge in the development of various innovative models.  
One example among the others, DPA-1~\citep{dpa1}, is a large-scale pre-trained model with improved attention architecture over DeepPot~\citep{wang2018deepmd}. Similarly, JMP~\citep{jmp} leverages diverse molecular systems of different types as a joint pre-training model. MACE~\citep{mace}, on the other hand, introduces higher-order message passing as a complete basis of many-body atomic interactions. Collectively, these models represent significant strides in the evolution of NNIP-driven atomistic simulations~\citep{escaip,liaoequiformerv2,gasteiger2021gemnet,nequip,schnet,smith2017ani,chgnet, alignn, mattersim}.

Nevertheless, the balance between computational efficiency and accuracy is key to practical applications of these models. On one side, efficient yet less expressive models are more suitable for tasks that weigh computational cost more than accuracy. On the other side, less efficient yet expressive models are more suitable for tasks where accuracy is more important.
Atomistic systems in 3D Euclidean space are invariant to Euclidean symmetries including rotation, translation and reflection. Thus, equivariant models are developed to respect these symmetries ~\citep{egnn,tfn}, and most of these expressive models by date~\citep{nequip,mace,liaoequiformerv2}  are based on the spherical harmonics~\citep{e3nn}, but calculating the tensor product of irreducible representations imposes expensive computational overheads. While another branch of work achieves rotation equivariance through building equivariant frames that can be either local or global~\citep{alphafold2,du2022se,leftnet,faenet}, and the main benefit of frame-based approaches is eliminating the necessity to involve the tensor product, thus greatly improving the computational efficiency. 

Herein, we propose a local-frame-based equivariant atomistic model, named AlphaNet, for accurate yet highly efficient atomistic simulations. Building on the success of frame-based NNIPs in small atomistic systems, we introduce an additional rotary position embedding to enhance the frame transition and temporal connection for multi-scale modeling. Extensive quantitative benchmarks demonstrate that AlphaNet excels at accuracy, efficiency and scalability compared with the state-of-the-art (SOTA) NNIP models on a variety of datasets, including formate decomposition, defected graphenes, zeolites, Open Catalyst 2020 (OC20), and  Matbench discovery WBM test set, which cover various types of interatomic interactions across a rich set of systems and phenomena, ranging from molecules and bulks to surfaces, from gas-phase reactions and crystal stability to surface catalysis.
\section{Results}

We systematically assess the capabilities of AlphaNet by examining its prediction accuracy, scalability with respect to model and dataset sizes, as well as computational efficiency in terms of inference speed and memory usage. Initially, we validate atomic-level prediction accuracy across five diverse chemical systems, simultaneously demonstrating AlphaNet's scalability concerning dataset and system sizes. Subsequently, we quantify computational efficiency by benchmarking both the memory footprint and inference costs of the model. Comprehensive details regarding datasets, comparative methodologies, and hyperparameter configurations are provided in~\Cref{sec:exp_details}.

\subsection{AlphaNet achieving high accuracy across all datasets}

We first consider the formate decomposition dataset, which serves as a representative example of catalytic surface reactions, specifically focuses on the dehydrogenation of formate (HCOO* → H* + CO$_2$) on a Cu $\langle 110 \rangle$ surface. For this dataset, AlphaNet achieves a mean absolute error (MAE) of 45.5 meV/\AA for force and 0.23 meV/atom for energy, compared to NequIP~\citep{nequip} 47.3 meV/\AA and 0.50 meV/atom, respectively, as shown in~\Cref{tab:model_performance}. These results underscore AlphaNet's ability to effectively capture the intricate nature of heterogeneous systems, which often involve both metallic and covalent bonding, along with complex charge transfer processes between the metal and adsorbed molecules. This high degree of accuracy highlights the model’s suitability for simulating catalytic reactions with multiple interaction types.
\vspace{-3mm}

\begin{table}[h]
\centering
\caption{Results on the Formate Decomposition and Defected Graphene datasets. Boldface indicates the best performance.}
\begin{tabular}{lccc}
\toprule
\textbf{Dataset} & \textbf{Metric} & \textbf{NequIP} & \textbf{AlphaNet} \\ 
\midrule
\multirow{2}{*}{\textbf{Formate Decomposition}} & Force MAE (meV/\AA) $\downarrow$ & 47.3 & \textbf{45.5} \\ 
                                    & Energy MAE (meV/atom) $\downarrow$ & 0.50 & \textbf{0.23} \\ 
\midrule
\multirow{2}{*}{\textbf{Defected Graphene}} & Force MAE (meV/\AA) $\downarrow$ & 60.2 & \textbf{19.4} \\ 
                                    & Energy MAE (meV/atom) $\downarrow$ & 1.9 & \textbf{1.2} \\ 
\bottomrule
\end{tabular}
\label{tab:model_performance}
\end{table}

\begin{figure}[h]
    \centering
    \begin{subfigure}[b]{0.45\textwidth}
        \centering
        \includegraphics[width=\textwidth]{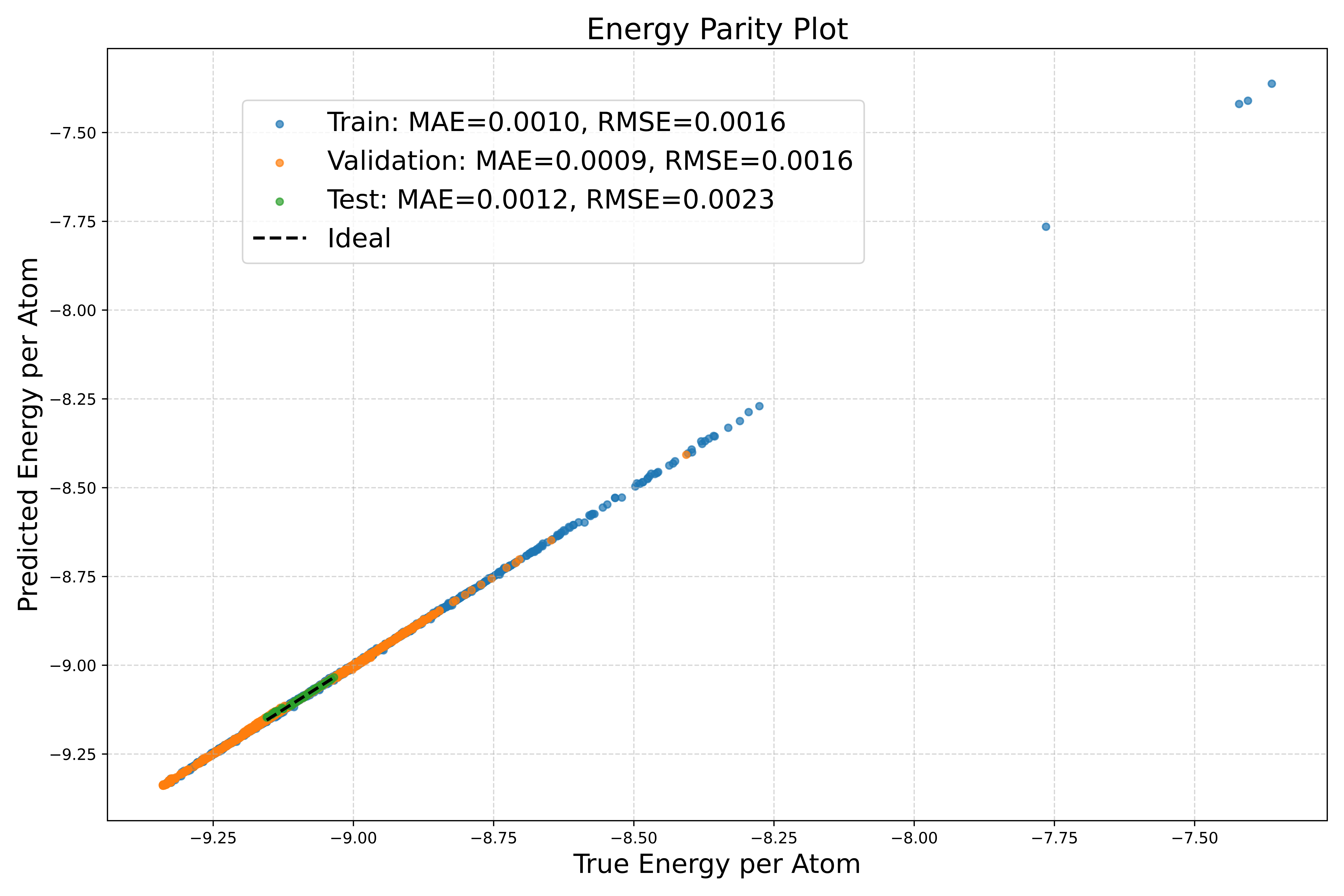}
        \caption{Parity plot for total energies obtained for the training (blue), validation (orange), and test (green) data sets.} 
        \label{fig:image1}
    \end{subfigure}
    \hfill
    \begin{subfigure}[b]{0.45\textwidth}
        \centering
        \includegraphics[width=\textwidth]{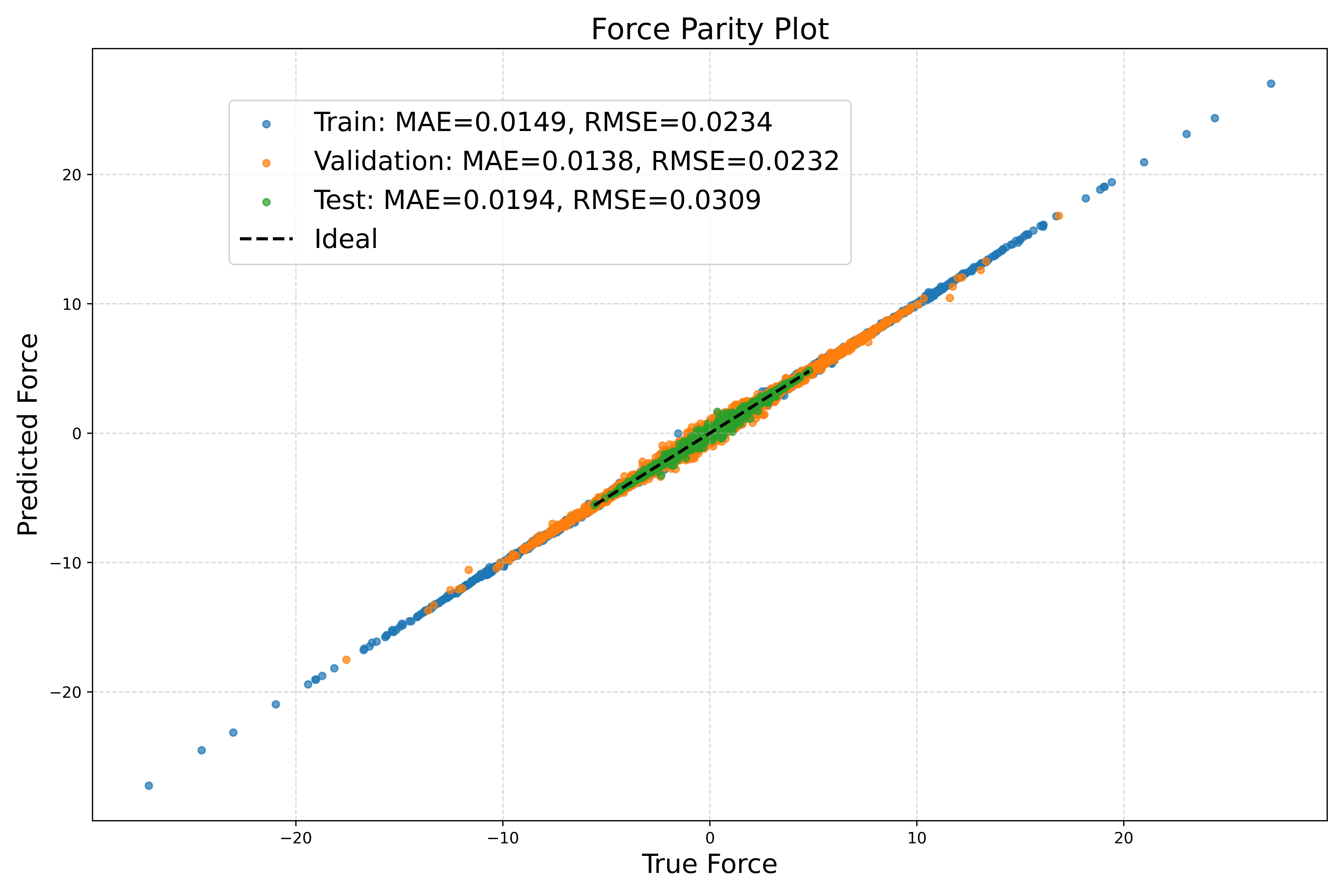}
        \caption{Parity plot for atomic forces obtained for the training (blue), validation (orange), and test (green) data sets.} 
        \label{fig:image2}
    \end{subfigure}
    

    \begin{subfigure}[b]{0.9\textwidth} 
        \centering
        \includegraphics[width=\textwidth]{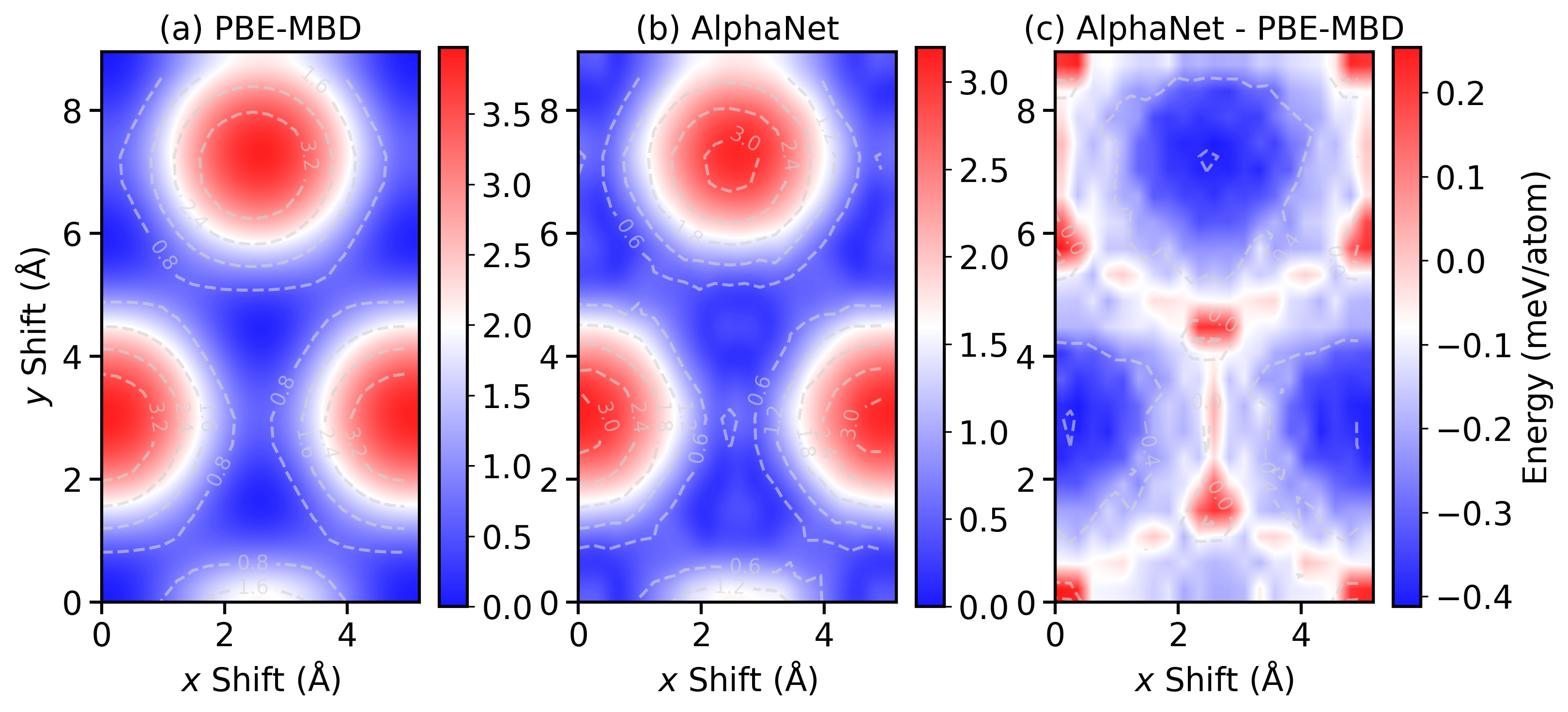}
        \caption{AlphaNet (middle) and DFT (left) sliding PES for bilayer graphene, and their difference (right).} 
        \label{fig:image3}
    \end{subfigure}

    \caption{Comparisons between AlphaNet predictions and DFT calculations on the Zeolite dataset.} 
    \label{fig:three_images}
\end{figure}

A further challenge for neural network interatomic potentials (NNIPs) is the accurate modeling of inter-layer sliding effects in layered materials such as defected graphene, where interlayer interactions are essential. On the defected graphene dataset, AlphaNet attains a force MAE of 32.0 meV/Å and an energy MAE of 1.7 meV/atom, significantly outperforming NequIP's 60.2 meV/Å and 1.9 meV/atom, respectively. Additionally, AlphaNet successfully reproduces the binding energy profile for AB-stacked bilayer graphene, benchmarked against PBE+MBD calculations, without explicitly incorporating long-range dispersion corrections. Deviations within the shallow sliding potential energy landscape are less than 0.4 meV/atom, highlighting AlphaNet's exceptional performance in modeling subtle interlayer forces and complex structural dynamics (see \Cref{fig:three_images}).

The third dataset evaluated is the zeolite dataset, consisting of 16 distinct zeolite types and comprising a total of 800,000 configurations. As summarized in \Cref{tab:performance_table}, AlphaNet demonstrates significantly improved performance compared to Deep Pot~\citep{wang2018deepmd} across nearly all zeolite configurations evaluated, exhibiting superior accuracy in both energy and force predictions. For example, in the ABWopt configuration, AlphaNet achieves an energy mean absolute error (MAE) of 12 meV, markedly lower than Deep Pot’s 90 meV. Similarly, AlphaNet attains a force MAE of 19 meV/\AA, clearly outperforming the 90 meV/\AA{} obtained by Deep Pot.

\begin{table*}[h!]
\centering
\caption{Results on the Zeolite dataset. Boldface indicates the best performance. (Unit for energy is meV/atom and unit for force is meV/\AA).}
\resizebox{\textwidth}{!}{%
\begin{tabular}{l*{8}{c}}
\toprule
 & \multicolumn{2}{c}{ABWopt} & \multicolumn{2}{c}{BCTopt} & \multicolumn{2}{c}{BPHopt} & \multicolumn{2}{c}{CANopt} \\
\cmidrule(lr){2-3}\cmidrule(lr){4-5}\cmidrule(lr){6-7}\cmidrule(lr){8-9}
\textbf{Metric} & Deep Pot & AlphaNet & Deep Pot & AlphaNet & Deep Pot & AlphaNet & Deep Pot & AlphaNet \\
\midrule
Energy MAE $\downarrow$  & 90  & \textbf{12} & 110 & \textbf{6.8} & 210 & \textbf{29} & 150 & \textbf{18} \\
Force MAE $\downarrow$   & 90  & \textbf{19} & 50 & \textbf{12}  & 60 & \textbf{16}  & 90  & \textbf{15} \\
\addlinespace[3pt]
 & \multicolumn{2}{c}{EDIopt} & \multicolumn{2}{c}{FERopt} & \multicolumn{2}{c}{GISopt} & \multicolumn{2}{c}{JBWopt} \\
\cmidrule(lr){2-3}\cmidrule(lr){4-5}\cmidrule(lr){6-7}\cmidrule(lr){8-9}
\textbf{Metric} & Deep Pot & AlphaNet & Deep Pot & AlphaNet & Deep Pot & AlphaNet & Deep Pot & AlphaNet \\
\midrule
Energy MAE $\downarrow$   & 40  & \textbf{8.7}  & 290 & \textbf{43} & 60  & \textbf{11}  & 150 & \textbf{10} \\
Force MAE $\downarrow$   & 50  & \textbf{13.5}   & 130 & \textbf{28}  & 50  & \textbf{11}  & 70  & \textbf{13} \\
\addlinespace[3pt]
 & \multicolumn{2}{c}{LOSopt} & \multicolumn{2}{c}{LTAopt} & \multicolumn{2}{c}{LTJopt} & \multicolumn{2}{c}{NATopt} \\
\cmidrule(lr){2-3}\cmidrule(lr){4-5}\cmidrule(lr){6-7}\cmidrule(lr){8-9}
\textbf{Metric} & Deep Pot & AlphaNet & Deep Pot & AlphaNet & Deep Pot & AlphaNet & Deep Pot & AlphaNet \\
\midrule
Energy MAE $\downarrow$   & 110 & \textbf{21}  & 150 & \textbf{19}  & 70  & \textbf{15}  & 210 & \textbf{16} \\
Force MAE $\downarrow$   & 70  & \textbf{12}  & 64  & \textbf{12}  & 110 & \textbf{11}  & 110 & \textbf{15} \\
\addlinespace[3pt]
 & \multicolumn{2}{c}{PARopt} & \multicolumn{2}{c}{PHIopt} & \multicolumn{2}{c}{SODopt} & \multicolumn{2}{c}{THOopt} \\
\cmidrule(lr){2-3}\cmidrule(lr){4-5}\cmidrule(lr){6-7}\cmidrule(lr){8-9}
\textbf{Metric} & Deep Pot & AlphaNet & Deep Pot & AlphaNet & Deep Pot & AlphaNet & Deep Pot & AlphaNet \\
\midrule
Energy MAE $\downarrow$   & 90  & \textbf{18}  & 60  & \textbf{20}  & 200 & \textbf{9.5}  & 160 & \textbf{17} \\
Force MAE $\downarrow$    & 70  & \textbf{24}  & 120 & \textbf{14}  & 110 & \textbf{12}  & 60  & \textbf{15} \\
\bottomrule
\end{tabular}%
}
\label{tab:performance_table}
\end{table*}

\begin{table}[h]
\centering
\caption{Results on the OC20 validation set. ``full'' denotes models trained on the full OC20 training set, others are trained on the 2M subset. $\lambda_E$ refers to the weight of the energy loss, the weight of the force loss $\lambda_F$ is set to $100$. Boldface indicates the best performance.}
\begin{tabular}{lcccc}
\toprule
\textbf{Model} & \textbf{Parameters $\downarrow$}& \textbf{Energy MAE (eV) $\downarrow$} & \textbf{Force MAE (eV/\AA) $\downarrow$}\\
\midrule
SchNet (full) & 9.1M & 0.54 & 0.55 \\
DimeNet++ (full) & 10.1M & 0.53  & 0.048\\
GemNet & 38M & 0.29 & 0.026\\
eSCN & 51M & 0.28 & 0.021 \\
EScAIP-Medium & 146M & 0.25 & \textbf{0.019}\\
EquiformerV2 ($\lambda_E=2$) & 146M & 0.28 & 0.022\\
\hline
AlphaNet ($\lambda_E=4$) & \textbf{6.1M} & 0.25 & 0.040 \\
AlphaNet ($\lambda_E=20$) & \textbf{6.1M} & \textbf{0.24} & 0.062 \\
\bottomrule
\end{tabular}
\label{tab:OC}
\end{table}

In addressing the next challenge, we consider the Open Catalyst Project OC20 dataset~\citep{OC}, a comprehensive collection of crystal structures specifically curated for training neural network interatomic potential (NNIP) models in catalysis applications. Our study focuses on the OC2M subset and evaluates performance on the Structure-to-Energy-and-Force (S2EF) task. We benchmark our proposed model, AlphaNet, against established state-of-the-art methods, including EquiformerV2~\citep{liaoequiformerv2}, EScAIP~\citep{escaip}, eSCN~\citep{escn}, GemNet-OC~\citep{gemnet}, SchNet~\citep{schnet}, and DimeNet++~\citep{dimenet}. AlphaNet, trained on the 2M subset for 2,000,000 steps, achieves remarkable accuracy in energy prediction, yielding a mean absolute error (MAE) of 0.24 eV. This accuracy is on par with larger-scale models such as EquiformerV2 and EScAIP. In contrast, models like SchNet and DimeNet++, even when trained on the full dataset, demonstrate inferior performance, with MAE values exceeding 0.35 eV. However, AlphaNet, trained exclusively on the smaller 2M subset, does not surpass the larger-scale models (EquiformerV2 and EScAIP) in force prediction tasks. This discrepancy likely arises from the inherent conservativeness of the model architectures, which enforce consistency by computing forces strictly as negative gradients of the predicted energies ($-\nabla_x E(x)$).

\begin{table}[h!]
\centering
\caption{Results on the Matbench Discovery WBM unique prototype dataset. Boldface indicates the best performance. }
\label{tab:matbench}
\begin{tabular}{lcccccccc}
\toprule
\textbf{Model} & \textbf{F1 $\uparrow$} & \textbf{DAF $\uparrow$} & \textbf{Prec $\uparrow$} & \textbf{Acc $\uparrow$} & \textbf{MAE $\downarrow$} & \textbf{R2 $\uparrow$} & \textbf{Size $\downarrow$} & \textbf{Type} \\
\midrule
CHGNet & 0.613 & 3.361 & 0.514 & 0.851 & 0.063 & 0.689 & 413k & EFSGM \\
MACE-MP-0 & 0.669 & 3.777 & 0.577 & 0.878 & 0.057 & 0.697 & 4.69M & EFSG \\
GRACE-2L-MPtrj & 0.691 & 4.163 & 0.636 & 0.896 & 0.052 & 0.741 & 15.3M & EFSG \\
SevenNet-0 & 0.724 & 4.252 & 0.65 & 0.904 & 0.048 & 0.75 & 842k & EFSG \\
SevenNet-l3i5 & 0.76 & 4.629 & 0.708 & 0.92 & 0.044 & 0.776 & 1.17M & EFSG \\
ORB MPtrj & 0.765 & 4.702 & 0.719 & 0.922 & 0.045 & 0.756 & 25.2M & EFSD \\
DPA3-v1-MPtrj & 0.765 & 4.654 & 0.711 & 0.921 & 0.042 & 0.798 & 3.37M & EFSG \\
eqV2 S DeNS & 0.815 & 5.042 & 0.771 & 0.941& 0.036 & 0.788 & 31.2M & EFSD \\
EScAIP & 0.782 & 5.634 & 0.712 & 0.939& 0.038 & 0.783 & 45M & EFSD \\ \hline
AlphaNet-L & 0.799 & 4.863 & 0.743 & 0.933 & 0.041 & 0.745 & 16.2M & EFSG \\
AlphaNet-S*\tablefootnote{* means not submitted yet} & 0.808 & 4.915 & 0.751 & 0.935 & 0.037 & 0.796 & 4.5M & EFSG \\
\bottomrule
\end{tabular}
\end{table}

Lastly, we validate the effectiveness of AlphaNet using the established Matbench discovery WBM test set~\citep{matbench}. AlphaNet demonstrates an excellent balance between predictive accuracy and physical consistency on the WBM benchmark (Table~\ref{tab:matbench}). Achieving an F1 score of 0.799 and a DAF score of 4.863, AlphaNet attains near state-of-the-art classification performance while utilizing an \textit{EFSG-type conservative force field} to rigorously enforce physical constraints. Notably, despite its compact architecture (16.2M parameters), AlphaNet-L surpasses the larger \textit{eqV2 S DeNS} model (31.2M parameters) in regression tasks, yielding an improved coefficient of determination ($R^2 = 0.742$) and lower prediction error (mean absolute error, $\text{MAE} = 0.039$).  There is also a new small size model(AlphaNet-S) with a different RBF function which performs even better. This exceptional combination of computational efficiency and physics-informed design positions AlphaNet as a powerful neural network interatomic potential (NNIP) model, ideally suited for scalable materials modeling.

\subsection{AlphaNet scaling effectively  with model, data and system sizes}

\begin{figure}[h]
    \centering
    \begin{subfigure}{0.48\textwidth} 
        \includegraphics[width=\linewidth]{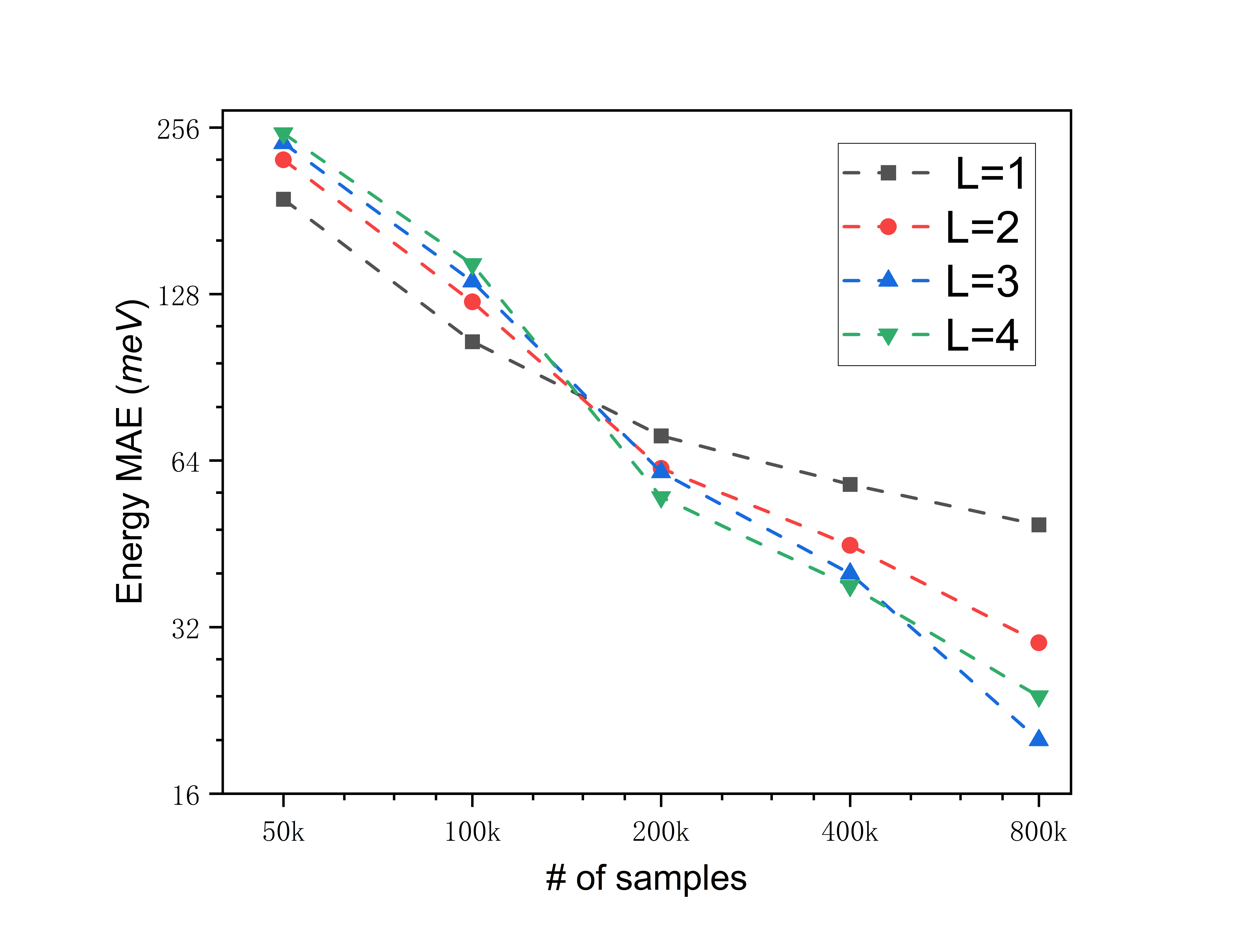}
        \caption{Energy MAE (meV) vs Data Size}
        \label{fig:energy_mae}
    \end{subfigure}
    \begin{subfigure}{0.48\textwidth} 
        \includegraphics[width=\linewidth]{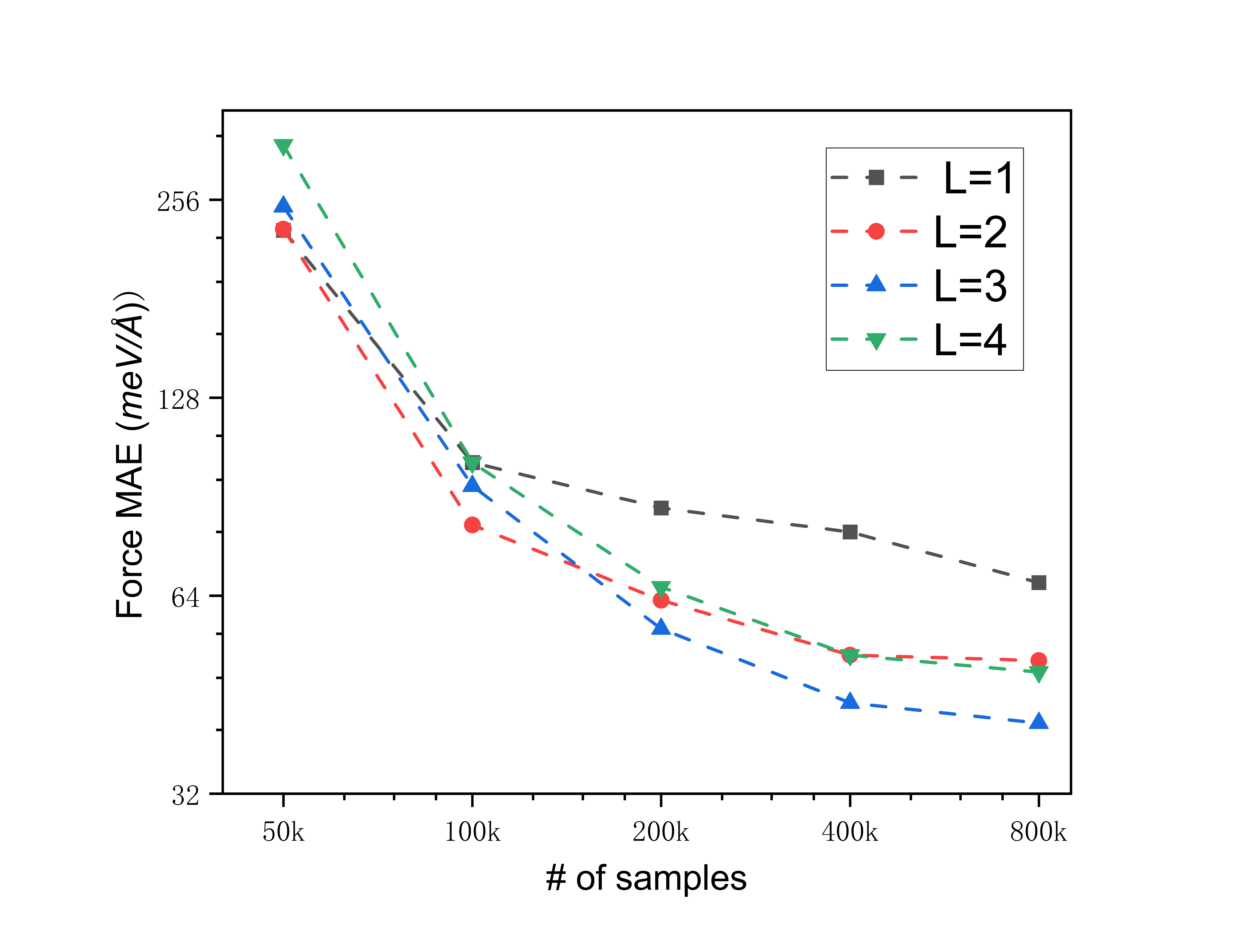}
        \caption{Force MAE (meV/\AA) vs Data Size}
        \label{fig:force_mae}
    \end{subfigure}
    \caption{Mean absolute errors of energy and force predictions of models with $L$ number of layers as a function of training data size on the zeolite dataset.} 
    \label{fig:mae_vs_data_size}
\end{figure}

In addition to demonstrating robust performance on large datasets and extensive systems, we systematically evaluate how model size and dataset scale influence performance using the zeolite dataset. As depicted in \Cref{fig:mae_vs_data_size}, increasing the training dataset size from 50k to 800k samples notably improves prediction accuracy. Specifically, the mean absolute error (MAE) for energy predictions decreases significantly from 220 meV at 50k samples to 41 meV at 800k samples. Similarly, the force prediction MAE reduces markedly from 241 meV/Å at 5k samples to 54 meV/Å at 80k samples. Furthermore, deeper neural networks exhibit accelerated performance improvements as dataset size increases, underscoring AlphaNet's scalability with respect to both dataset and model size.

\subsection{AlphaNet demonstrating enhanced speed-accuracy trade-off}
The primary advantage of NNIP models over density functional theory (DFT)-calculated energies and forces is their substantially faster inference speed. However, increasing the number of parameters in an NNIP model to enhance its predictive accuracy typically results in slower inference speeds. To systematically assess inference speed across various NNIPs, we measure the average forward-pass computation time using 2,000 zeolite structures, processed with a batch size of 10. Additionally, we evaluate inference performance across systems of varying sizes by systematically expanding their unit cells. While we strive to maintain a consistent parameter count across models, minor variations persist due to differing internal architectures and components, as detailed in~\Cref{sec:baseline}.

\begin{figure}[h]
   \begin{subfigure}{0.48\textwidth} 
    \raisebox{-0.5em}{\includegraphics[width=\linewidth]{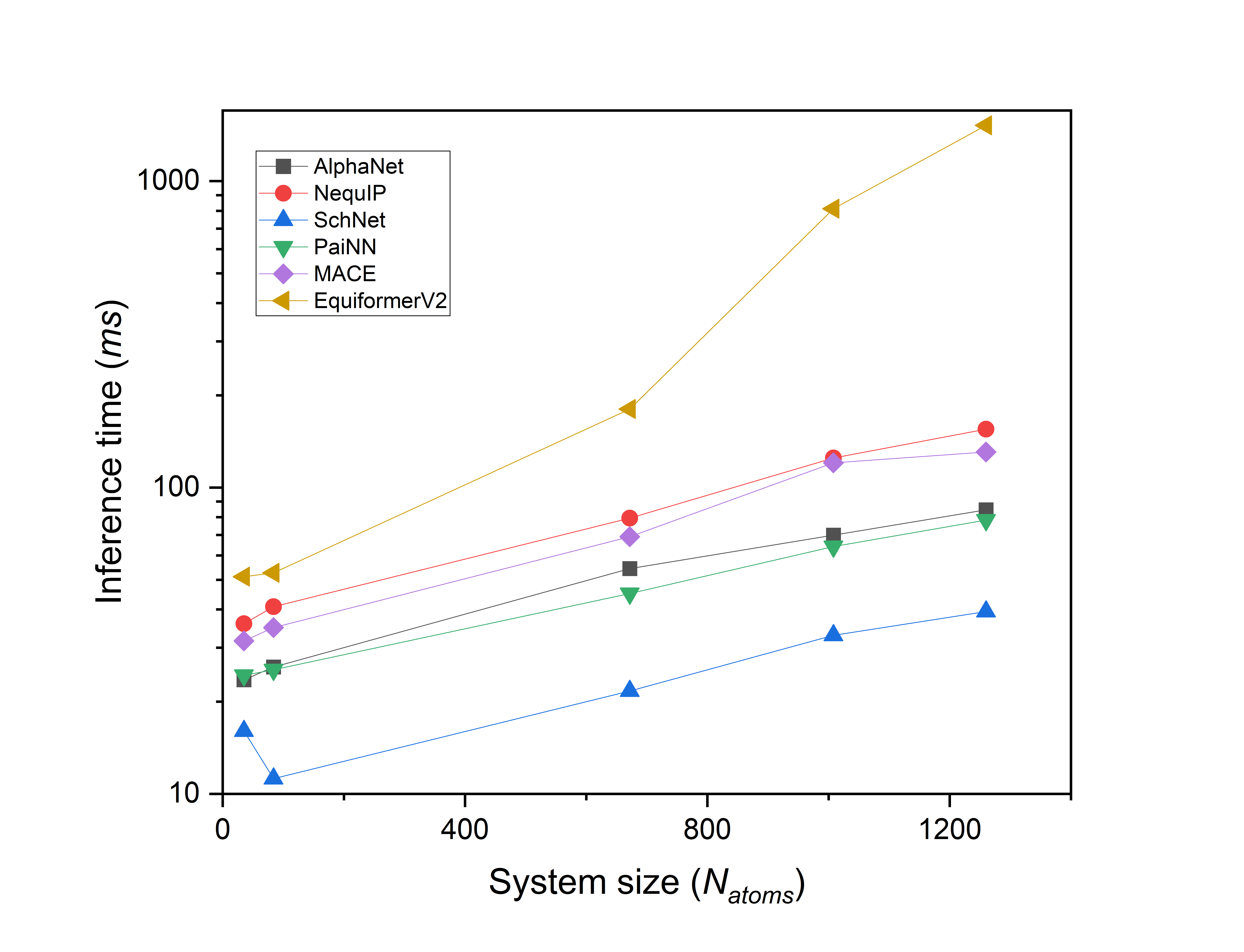}} 
    \caption{Inference speed as a function of system size on the zeolite dataset. }
    \label{fig:speed1}
   \end{subfigure}
   \begin{subfigure}{0.4\textwidth}
     \includegraphics[width=\linewidth, height=0.8\linewidth]{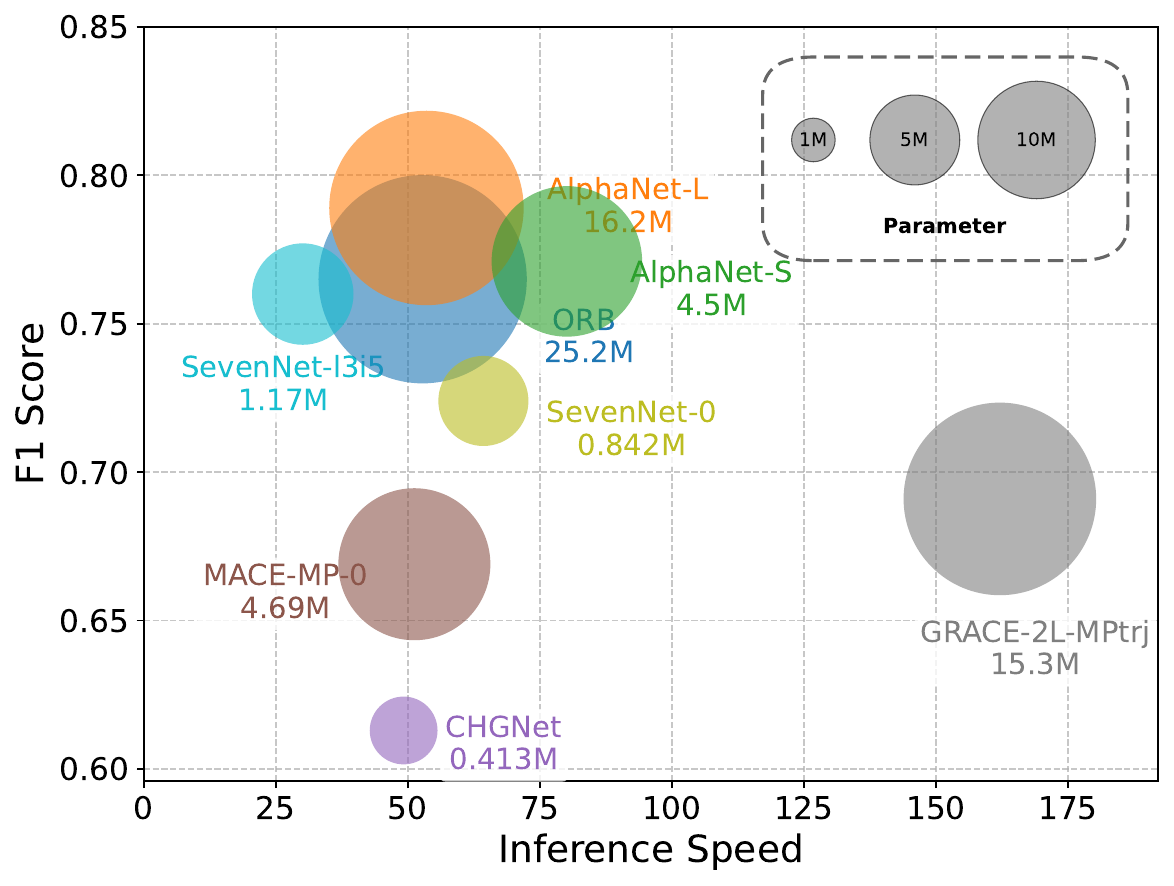}
     \caption{Inference speed vs performance on Matbench Discovery.}
     \label{fig:speed2}
    \end{subfigure}
    \vspace{-3mm}
    \caption{Comparisons of inference speed among various NNIP models. (a). The x-axis represents the number of atoms in the system, and the y-axis shows the average time required to predict energy and forces for systems in batches of 10, across a total of 200 batches on the Zeolite dataset. (b). All models are accessible and compliant with the Matbench Discovery leaderboard. The speed is calculated as the number of energy evaluation per second.}
\label{fig:speed}
\end{figure}

As shown in \Cref{fig:speed}(a), AlphaNet exhibits a clear computational advantage over other NNIP models, characterized by a notably slower increase in computational time with growing system size. This efficient scaling behavior makes AlphaNet particularly well-suited for simulations involving larger systems. Furthermore, we assess the inference speed of pre-trained models featured on the Matbench Discovery leaderboard~\citep{matbench}. To ensure a fair comparison, we select compliant, open-source models that are freely accessible without additional authorization. The average energy prediction speeds of these models are presented in \Cref{fig:speed}(b). The evaluated models include AlphaNet, ORB-v2~\citep{orb}, SevenNet~\citep{7net}, GRACE~\citep{grace}, MACE~\citep{mace}, and CHGNet~\citep{chgnet}. Among these, GRACE implemented using TensorFlow demonstrates significantly faster inference compared to the other models, which are built on PyTorch, and our model, AlphaNet-S, comes in the second place. In particular, AlphaNet-L exhibits an inference speed comparable to models possessing approximately ten times fewer parameters, such as SevenNet-l3i5. This finding indicates that AlphaNet maintains competitive computational efficiency despite its relatively larger size, raising an interesting prospect for investigating its potential performance with reduced parameters in future studies.

\begin{table*}[h]
\centering
\label{tab:OC_eff}
\caption{Efficiency evaluated on the OC20 dataset, using a single A100 40GB GPU. Conserv. denotes energy conservation. Boldface indicates the best performance.} 
\begin{adjustbox}{width=.95\linewidth}
\begin{tabular}{ccccccc}
\toprule
\textbf{Models} & \textbf{Parameters}  $\downarrow$ & \textbf{\makecell[c]{Training\\Speed\\(Sample/Sec) $\uparrow$}} & \textbf{\makecell[c]{Training\\Memory\\(GB/Sample) $\downarrow$}} & \textbf{\makecell[c]{Inference\\Speed\\(Sample/Sec) $\uparrow$}} & \textbf{\makecell[c]{Inference\\Memory\\(GB/Sample) $\downarrow$}} & \textbf{Conserv.} \\
\midrule
SchNet & 9.1M & 39 & 2.5 & 112.1 & 1.7 & \checkmark \\
DimeNet++ & 10.1M & 29 & 3.6 & 92 & 2.5 & \checkmark \\
GemNet-OC & 38M & 16.22 & 1.63 & 39.2 & 1.61 & $\times$ \\
eSCN & 51M & 8.5 & 1.75 & 25.2 & 1.74 & $\times$ \\
EScAIP & 83M & 37.8 & \textbf{1.23} & 111.8 & \textbf{1.09} & $\times$ \\
EquiformerV2 & 31M & 15.7 & 1.63 & 36.72 & 1.61 & $\times$ \\
\hline
AlphaNet & \textbf{6.1M} & \textbf{41.2} & 2.2 & \textbf{120.1} & 1.5 & $\checkmark$ \\
\bottomrule
\end{tabular}
\end{adjustbox}
\end{table*}

The final efficiency evaluation is conducted using the OC20 dataset, with results summarized in \Cref{tab:OC_eff}. Notably, EScAIP demonstrates outstanding efficiency in both inference speed and memory usage, even given its substantial model size. AlphaNet exhibits a marginally faster inference speed; however, it incurs higher memory consumption compared to EScAIP. It is important to emphasize that models such as AlphaNet, SchNet, and DimeNet++ are conservative in their force calculations, inherently relying on PyTorch's autograd functionality, thus necessitating additional computational overhead and consequently increasing inference time and memory usage.

\section{Discussion}
AlphaNet represents a significant advance in scalable neural network interatomic potentials, achieving exceptional accuracy while maintaining computational efficiency. Its success stems from a novel local-frame-based architecture that scalarizes geometric information within local frames and effectively aggregates these frames to produce features sensitive to both local and global chemical environments. Leveraging this architecture, AlphaNet adeptly captures complex atomic interactions across diverse chemical bonding types, including metallic, covalent, ionic, and long-range interactions, demonstrating versatility and broad applicability.

One of AlphaNet's key strengths is its exceptional scalability. The model consistently maintains high performance across varying dataset size and complexity of atomic systems. Remarkably, AlphaNet achieves robust accuracy while utilizing fewer parameters compared to conventional approaches, enhancing its computational efficiency. This capability to effectively manage large-scale datasets without compromising performance positions AlphaNet as a highly promising framework for extensive simulations and practical real-world applications.

In future work, we plan to extend the model's capabilities to account for additional interaction types, notably hydrogen bonding, which plays a critical role across diverse chemical and biological systems. Another important direction is enhancing the interpretability of the proposed model. Although our current implementation demonstrates strong predictive performance, its applicability and reliability in practical scenarios can be significantly improved by achieving deeper insight into how the model identifies, processes, and leverages atomic-level interactions.

Additionally, addressing the efficient use of GPU memory during training presents a long-term challenge, particularly for large systems exceeding 1,000 atoms—even when leveraging parallelization across multiple GPUs. This limitation, common among most message-passing neural networks, must be overcome to facilitate scalability toward larger and more complex systems. Future research will prioritize memory optimization strategies to surmount this bottleneck, ensuring broader applicability and enabling analysis of increasingly intricate molecular systems.

\section{Methodology}


\textbf{Neural Network Interatomic Potentials.} A single state of a molecular system can be described by a set of atom types $h \in \mathbb{R}^{n\times d}$ and atomic positions $x \in \mathbb{R}^{n\times 3}$. In molecular dynamics, we are interested in learning a function $f: \mathbb{R}^{n\times d} \times \mathbb{R}^{n\times 3} \rightarrow \mathbb{R}$ which predicts the energy of the current state of the molecular system readily to be used for simulation. The force can be further calculated by $-\nabla_x f(x)$, thus often referred as to neural network interatomic potentials (NNIPs). 

\textbf{Message-passing neural networks.} One common way to model molecular systems is through message-passing neural networks, which is a general neural architecture that leverages both permutation equivariance and locality in graph-structured data. Specifically, there are two essential operations in message-passing neural networks, message calculation and message update. The message calculation operation encodes structured messages from local environments $m_i = \oplus_{j \in \mathcal{N}(i)} f_m(f_h(x_i), h(x_j), e_{ij})$ where $i$ and $j$ denotes the index of different nodes, $e_{ij}$ denotes optional edge features and $f_m$ and $f_h$ denotes neural networks that encode both edge and node features; The message update aggregates the incoming messages from the neighbors $m_i = f_u(m_i)$, where $f_u$ is a neural network. 

\textbf{Equivariant message-passing neural networks.} Equivariance is another crucial property to building MLFFs as molecular systems are invariant to the Euclidean group. A function $f$ is said to be equivariant with respect to certain group $G$ if $f \circ g (x) = g \circ f(x)$, $\forall x \in X,\, g\in G$. One natural property of message-passing neural networks is the permutation equivariance such that changing the order of the input set of nodes do not result in difference output for the same node as it only depends on the neighbors which are unchanged under permutation. In addition to permutation, we need to consider the special Euclidean group in 3D, SE(3), which includes rotation and translation operations. We do so by leveraging local frames~\citep{du2022se}. Specifically, we build a set of equivariant and complete frames and scalarize the geometric quantity which does not break SE(3)-equivariance under nonlinear neural network encodings.

To build equivariant and complete frames from atomic positions, we consider the following edge-based frames following~\citep{du2022se}:
\[
\mathcal{F}_{ij} = \left( \hat{e}^1_{ij}, \hat{e}^2_{ij}, \hat{e}^3_{ij} \right) = \left( \frac{x_i - x_j}{\| x_i - x_j \|}, \ \frac{(x_i - \overline{x}) \times (x_j - \overline{x})}{\| (x_i - \overline{x}) \times (x_j - \overline{x}) \|}, \ \frac{(x_i - x_j) \times \left( (x_i - \overline{x}) \times (x_j - \overline{x}) \right)}{\| (x_i - x_j) \times \left( (x_i - \overline{x}) \times (x_j - \overline{x}) \right) \|} \right)
\]
To scalarize geometric quantity such as a vector $v$, we take the inner product between it and the frames which results in a set of invariant scalars and can be inverted via a tensorization operation
\[
\text{Scalarize}(v, \mathcal{F}_{ij}) = (s^1_{ij}, s^2_{ij}, s^3_{ij}) =  (e^1_{ij} \cdot v, e^2_{ij} \cdot v, e^3_{ij} \cdot v)
\]
\[
\text{Tensorize}(s_{ij}, \mathcal{F}_{ij}) = s^1_{ij} e^1_{ij} + s^2_{ij} e^2_{ij} + s^3_{ij} e^3_{ij}
\]

\textbf{Efficient and expressive equivariant message-passing neural networks.} Following \citep{du2024new}, we leverage an efficient and expressive extension of the above equivariant message-passing neural networks based on local frames. We consider two modules to improve the expressiveness while maintaining the efficiency of the network. Specifically, we leverage a local structure encoding module which scalarizes not only features from local neighbors but also from overlapping neighbors along each edge, denoted as $A_{ij} := f_l(\text{Scalarize}(S_{i-j}, \mathcal{F}_{ij}))$, where $f_l$ denotes a neural network and $S_{i-j}$ represents overlapping neighbors of node $i$ and $j$. In addition, we use an additional frame transition module which encodes the alignment between each neighboring frame to improve the expressiveness. We realize this by incorporating vector-based message passing to amortize the cost where $\bold{m}_{ij}$ denotes equivariant edge features. This framework is highly efficient as it avoids expensive higher-body message-passing neural networks and higher-order tensor updates.
\[
\{h_i, x_i\} = \{f_u(\oplus_{j \in \mathcal{N}(i)} f_m(h(x_i), h(x_j), A_{ij} , e_{ij})), \oplus_{j \in \mathcal{N}(i)} f_{\bold{m}}(h(x_i), h(x_j), A_{ij}, e_{ij}) \bold{m}_{ij} \}
\]
\textbf{Rotary positional embedding as invariant frame transition.} Positional embedding are known to improve and stabilize the training of transformers~\citep{shaw2018self}. We introduce the elegant relationship between the commonly used Rotary Position Embedding (RoPE) and frame transition we use in our architecture~\citep{su2024roformer}. In fact the frame transition can be considered as a generalized rotary positional embedding for 3D equivariant frames. Note that the spirit of rotary position embedding is to design an embedding scheme for relative position in sequential data. 
To build the rotary positional embedding, ~\citep{su2024roformer} utilizes the relative index between nodes $x_i$ and $x_j$ of a one-dimensional sequence to find a function $g(x_i,x_j,i-j)$ such that
\[
\langle f_q(x_i,i),f_k(x_j,j)\rangle = g(x_i,x_j,i-j)
\]
where $f_q$ and $f_k$ denote the neural network embeddings of query and key inside an attention layer. Fortunately, there is an explicit way of building $g$ by multiplying $f_{\{q,k\}}(x_i,i)$ with the diagonal matrix expanded by rotation matrices~\citep{su2024roformer}:
\[
\begin{pmatrix}
\cos \theta_n , & -\sin \theta_n\\
\sin \theta_n , & \cos \theta_n
\end{pmatrix}
\]
where $\theta_n = 10000^{-2\frac{(n-1)}{d}}, \quad n \in [1, 2, \ldots, \frac{d}{2}]$, and $d$ is the dimension of features.
From a geometric point of view, the index $i$ of $x_i$ provides a one-dimensional position embedding such that the inner product $\langle f_q(x_i,i),f_k(x_j,j)\rangle$ only depend on the relative position $i-j$. In the three-dimensional world, we can utilize the frame transition matrices between $x_i$ and $x_j$ that plays a similar rule of the index difference $i-j$ in a one-dimensional sequence. In addition, we introduce the rotary positional embedding on invariant features $h$ and represent the rotation by the multiplication of complex numbers. We predict an additional complex number from our invariant features $h_i$ for each node $i$ and multiply it by the scalarized features $h_i$. Similar to the equivariant message passing in each layer of the network, the learned rotary embedding is also applied in each layer. 

\textbf{Temporal connection.} In addition to commonly adapted message passing in the spatial domain, we consider another dimension, temporal domain, which can be described by the depth of the neural network~\citep{chen2018neural}. As message-passing neural networks are local, we believe it is beneficial to integrate multi-scale information. Specifically, for each consecutive layer, we learn a kernel that linear transforms and aggregates features.
Given the invariant feature embeddings $h^{l_1}$ at layer $l_1$ and $h^{(l_2)}$ at layer $l_2$, the kernel is of the matrix product form: $M^{(l_{3,1,2})}$ which transforms $h^{(l_1)}$ and $h^{(l_2)}$ to a new feature $h^{(l_3)}$ and added as a residual into the next layer: 
\[h^{(l_3)} = \sum_{k_1=1}^{K_1} \biggl( \sum_{k_2=1}^{K_2} M^{(l_{3,1,2})}_{k_1, k_2} \cdot h^{(l_2)}_{k_2}\biggl) \cdot \, h^{(l_1)}_{k_1}
\]

\begin{figure}[h] 
    \centering
    \includegraphics[width=\linewidth]{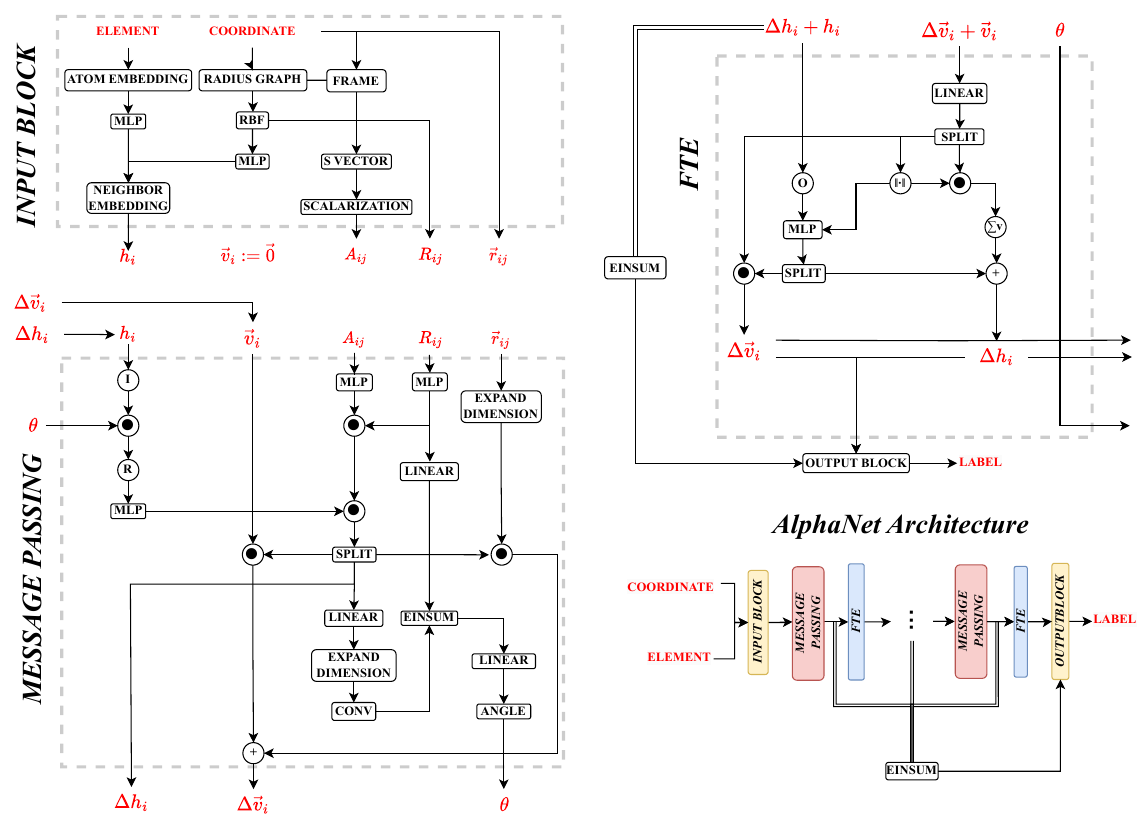} 
    \caption{Overview of AlphaNet framework. We first process the input atomic types and coordinates to scalarized features and frames and pass them further to a loop of message passing and frame transition layers, followed by an output block with a temporal connection to the final output. $I/R$ denotes imaginary and real number, $\|\bullet\|$ denotes norm, $\bullet$ denotes element-wise multiplication, and $\circ$ denotes vector scaling.} 
    \label{fig:model}  
\end{figure}

\bibliographystyle{abbrvnat} 
\bibliography{main.bib}

\newpage
\appendix
\begin{center}
{\Large \textbf{Supplementary Information for}
\\ AlphaNet: Scaling Up Local-frame-based Neural Network Interatomic Potentials}
\end{center}

\maketitle

\section{Local Structures, Completeness, and Expressive Power of GNNs}

Graph neural networks (GNNs) derive their \textbf{locality} from aggregating information within $k$-hop neighborhoods, typically limited to 1-hop interactions. While this design enables efficient learning, it inherently restricts their focus to local substructures such as trees, triangles, or cycles. These substructures define distinct \textbf{local isomorphism types}: (1) \textit{Tree isomorphism}, where subgraphs share identical hierarchical parent-child relationships; (2) \textit{Triangular isomorphism}, requiring matching cyclic triplets (e.g., molecular bond angles); and (3) \textit{Subgraph isomorphism}, preserving node features and connectivity patterns.

The expressive power of permutation-invariant GNNs is classically evaluated via the 1-Weisfeiler-Lehman (1-WL) test ~\citep{wl-test}. As established by ~\citep{wl2}, the 1-WL test is equivalent to a GNN's capacity to discriminate \textit{local subtree isomorphism}, measuring its ability to distinguish hierarchical neighborhood structures. However, this framework operates solely on 2D graph topologies, ignoring 3D geometric relationships (e.g., bond angles, torsional dihedrals), thereby failing to address molecular or material systems where spatial configurations are decisive.

 \textbf{Completeness} in GNNs requires a transformation function $ f $ to satisfy $ f(X) = f(Y) \iff X \text{ and } Y \text{ are isomorphic} $. We can extends completeness to \textbf{local geometric isomorphism}, where isomorphism is defined under 3D geometric constraints (distances, angles, chiralities). Concurrently, methods encoding 3D geometric invariants ~\citep{wl2} have demonstrated enhanced expressivity, enabling differentiation of graphs with distinct local atomic environments beyond WL limitations. Similar to LeftNet~\citep{leftnet}, AlphaNet is able to differentiate local structures beyond tree isomorphism by defining the 3D structure weights.

\section{Experimental Details}
\label{sec:exp_details}

\subsection{Datasets}
\label{sec:dataset}


\textbf{The Defected Bilayer Graphene Dataset~\citep{graphenedata}} consists of reference structures specifically designed to train and validate machine learning potentials (MLPs). It includes three bilayer systems: V0V0 (pristine), V0V1 (single vacancy on the top layer), and V1V1 (single vacancy in both layers). The dataset comprises single-point DFT (PBE+MBD) energies and atomic forces calculated for various configurations, including different interlayer distances, stacking modes, and manually deformed structures. Additionally, snapshot configurations from MD simulations at different temperatures were included. The data were split into training, validation, and test sets, containing 3988, 4467, and 200 structures, respectively. Farthest point sampling (FPS) and principal component analysis (PCA) were used to ensure a representative and balanced division. 

\textbf{The Formate Decomposition on Cu Dataset~\citep{nequip}} consists of configurations related to the decomposition process of formate on a Cu surface, specifically involving the cleavage of the C-H bond. The dataset includes initial, intermediate, and final states, such as monodentate and bidentate formate on Cu <110>, as well as the final state with an H ad-atom and desorbed CO2 in the gas phase. Nudged elastic band (NEB) method~\citep{neb} was used to generate an initial reaction path for the C-H bond breaking, followed by 12 short ab initio molecular dynamics (AIMD) simulations. These simulations were performed using the CP2K code, resulting in a total of 6855 DFT structures with a time step of 0.5 fs and 500 steps per trajectory. AlphaNet was trained on 2500 uniformly sampled structures from the full dataset, with a validation set of 250 structures and the mean absolute error evaluated on the remaining structures. 

\textbf{The Zeolite Dataset} comprises essential reference structures designed for various applications such as catalysis, adsorption, and separation. Zeolites are highly valued porous materials that have been the subject of extensive scientific research due to their broad applicability. The dataset includes 16 distinct zeolites, with atomic-level trajectories obtained through \textit{Ab Initio Molecular Dynamics} (AIMD)~\citep{aimd} simulations conducted at 2000 K using the Vienna \textit{Ab Initio Simulation Package} (VASP)~\citep{vasp,vasp2}. For each zeolite, 80,000 snapshots were extracted, and the corresponding energies and atomic forces were calculated for every configuration. The dataset was randomly divided into training, validation, and test sets in a 6:2:2 ratio, comprising 48,000, 16,000, and 16,000 structures, respectively.

\textbf{The OC2M Dataset~\citep{OC}} is a large-scale subset of the OC20 Dataset, specifically designed for training and evaluating machine learning models in the context of interatomic potentials. This dataset includes approximately 2 million atomic configurations in the training set, with validation and test sets containing 4 splits, which is In Domain (ID), out-of-domain adsorbate (OOD adsorbate), out-of-domain catalyst (OOD catalyst), and both of unseen adsorbate and unseen catalyst (OOD both). Each of them contains around 1 million structures. It encompasses configurations involving 56 different chemical elements, ensuring a broad and diverse representation of atomic environments. These configurations were generated using density functional theory (DFT) calculations, with a focus on a wide range of materials science applications, including bulk materials, surfaces, and defect structures. Originating from the comprehensive OC20 Dataset, the OC2M subset provides accurate total energies and atomic forces for each configuration. This extensive dataset is particularly valuable for developing and testing machine learning models that require a diverse and large-scale collection of training examples to achieve robust generalization across different types of atomic interactions.

\textbf{The MPTraj Dataset and Matbench Discovery Test~\citep{matbench}}. MPTraj consists of 1,580,395 structures which are frames of the DFT relaxations performed on all 154,719 Materials Project materials~\citep{chgnet}. Matbench Discovery is a test task for predicting the stablity of 256,963 materials, which is also known as WBM test set~\citep{wbm}. In this case we train our model on the MPTraj dataset and test it on the WBM test set. 

\subsection{Hyperparamters}
\label{sec: Hyp}

\begin{table}[ht]
\centering
\small  
\caption{Hyperparameter configurations for five different datasets. Gra. denotes the defected graphene dataset, For. denotes the formate decomposition dataset, Zeo. denotes the zeolite dataset, OC2M denotes the OC2M dataset, MPTraj denotes the MPTraj dataset.}
\label{tab:hparams}
\begin{tabular}{@{}lccccc@{}}
\toprule
\textbf{Parameter} & \textbf{Gra.} & \textbf{For.} & \textbf{Zeo.} & \textbf{OC2M} & \textbf{MPTraj} \\ 
\midrule
Learning rate (lr)        & 5e-4 & 5e-4 & 5e-4 & 1e-4 & 1e-4 \\
grad\_clip                & 0.5 & 0.5 & 0.5 & 0.5 & 0.5 \\
Scheduler                & Cos. & Cos. & Cos. & Cos. & Cos. \\ 
Epochs                   & 150 & 150 & 100 & 130 & 200 \\
Loss func.               & MAE & MAE & MAE & MAE & MAE \\
Batch size (per device)  & 24 & 12 & 96 & 4 & 12 \\ 
\midrule
Num layers               & 3 & 3 & 3 & 4 & 6 \\
Heads                    & 16 & 16 & 16 & 16 & 24 \\ 
Num basis                & 32 & 32 & 32 & 96 & 256 \\
Hidden channels          & 128 & 128 & 128 & 256 & 256 \\
Cutoff (\AA)             & 5 & 5 & 5 & 5 & 6 \\
\bottomrule
\end{tabular}
\end{table}

\subsection{Training cost on the OC2M dataset and MPTraj}
\begin{table*}[h]
\centering
\caption{Training cost for the OC2M and MPTraj datasets. The training memory here is determined by dividing the fixed memory capacity of a GPU by the batch size. }
\resizebox{0.95\linewidth}{!}{
\begin{tabular}{lcccc}
\toprule
\textbf{Dataset} & \textbf{Training Time (GPU Days) } & \textbf{Number of Parameters} & \textbf{Training Memory (GB per sample)}\\
\midrule
 OC2M & 361 & 6.1M & 5.1\\
 MPTraj & 243 & 16.2M & 10.1\\
\bottomrule
\end{tabular}
}
\label{tab:OC_sup}
\end{table*}

\subsection{Baseline Models}
\label{sec:baseline}

For computational modeling configurations, NequIP achieves optimal performance with tensor rank 2, feature size 32, and 5\,\AA cutoff in the formate decomposition dataset, while using tensor rank 1 with 7\,\AA cutoff for defective graphene studies. The zeolite dataset employs a fine-tuned DPA-2 potential via Deep Pot implementation (\citep{dpa2}). Models trained on OC20 leverage pre-trained SchNet/DimeNet++ architectures from \url{https://fair-chem.github.io/core/model_checkpoints.html}, with other results cross-validated against \citep{escaip} using publicly accessible validation sets. MPtrj benchmark data integrates results from \url{https://matbench-discovery.materialsproject.org/} and comparative analysis in \citep{escaip}. Speed testing on zeolite structures maintains uniform 2-layer architectures with 5\,\AA cutoff, where PaiNN/SchNet employ 120-atom basis and 16 RBF basis, MACE uses tensor order 0-1 (feature size 16), NequIP extends to tensor order 0-2 (feature size 16), AlphaNet configures 32 hidden channels with 12 attention heads, and EquiformerV2 implements sphere/edge channels 16 with 16-dimensional attention hidden states and 2-head mechanisms.

\section{Ablation Study}

\begin{table}[ht]
\centering
\caption{Ablation study on the impact of RoPE, temporal connections (Temp Connect), and frame transition encoding (FTE).}
\label{tab:ablation}
\begin{tabular}{lcc}
\toprule
\textbf{Configuration} & \textbf{Energy (meV/atom)} $\downarrow$ & \textbf{Force MAE (meV/\AA)} $\downarrow$ \\
\midrule
Full Model & 1.7 & 32.0 \\
\hline
w/o RoPE & 25.9 & 113.6 \\
w/o Temp Connect & 2.9 & 38.2 \\
w/o FTE & 2.8 & 45.3 \\
\bottomrule
\end{tabular}
\end{table}

We do ablation study on the Defected Graphene dataset, ~\Cref{tab:ablation} shows that all the component matters, but it is worth notice that our model can not get down the loss and performs bad with out RoPE, though this did not happen in our other tests. While shows that RoPE is important, some other aspects may need to be considered such as more efficient way of optimizing tensor network.

\section{Data Scaling and Dynamic loss weights}

The experimental datasets were generated using \textit{VASP}~\citep{vasp,vasp2}, where the energy per atom typically converges around $5\,\mathrm{eV}$ under our default training settings. These configurations demonstrated robust performance within this energy regime. However, when applied to systems with significantly higher energy magnitudes per atom and greater energy variance---such as those produced by \textit{CP2K}~\citep{cp2k} simulations---the original settings failed to achieve stable convergence. To address this challenge, we developed a dynamic optimization strategy: initially adjusting the energy-to-force loss weight ratio from $\lambda_E:\lambda_F = 4:100$ to $0.05:100$, then progressively increasing $\lambda_E$ while simultaneously decaying the learning rate. This coordinated adaptation proved critical for successful training in high-variance scenarios.

\end{document}